%% file: arxiv.tex
\documentclass[10pt,twocolumn,letterpaper]{article}

\usepackage{cvpr}              
\usepackage{multirow}
\usepackage[table,xcdraw]{xcolor}
\usepackage{booktabs}
\usepackage{float}
\input{preamble}
\definecolor{cvprblue}{rgb}{0.21,0.49,0.74}
\usepackage[pagebackref,breaklinks,colorlinks,allcolors=cvprblue]{hyperref}

\def\paperID{17197} 
\def\confName{CVPR}
\def\confYear{2026}

\title{Think-Then-Generate: Reasoning-Aware Text-to-Image Diffusion \\ with LLM Encoders
}
\author{Siqi Kou$^{1*}$, Jiachun Jin$^{1*}$, Zetong Zhou$^{1*}$, Ye Ma$^{2}$,
Yugang Wang$^{1}$, Quan Chen$^{2}$, Peng Jiang$^{2}$, \\ Xiao Yang$^{3}$, Jun Zhu$^{3}$, Kai Yu$^{1}$, Zhijie Deng$^{1\dagger}$ \\
        \textsuperscript{1}Shanghai Jiao Tong University\; \textsuperscript{2}Kuaishou Technology\; \textsuperscript{3}Tsinghua University\\
\href{https://github.com/zhijie-group/Think-Then-Generate}{https://github.com/zhijie-group/think-then-generate}
}

\begin{document}
\maketitle
\renewcommand{\thefootnote}{}
\footnote{\textsuperscript{$*$}Equal contribution. Work done during Jiachun and Siqi’s internships at Kuaishou Technology.}
\footnote{\textsuperscript{$\dagger$}Corresponding author.}
\renewcommand{\thefootnote}{\arabic{footnote}}
\setcounter{footnote}{0}

\input{sec/0_abstract}    
\input{sec/1_intro}
\input{sec/2_preliminary}
\input{sec/3_method}
\input{sec/4_experiment}
\input{sec/5_related}
\input{sec/6_conclusion}
{
    \small
    \bibliographystyle{ieeenat_fullname}
    \bibliography{main}
}

\input{sec/X_suppl}

\end{document}

%% file: sec/0_abstract.tex
\vspace{-6pt}
\begin{abstract}
Recent progress in text-to-image (T2I) diffusion models (DMs) has enabled high-quality visual synthesis from diverse textual prompts. 
Yet, most existing T2I DMs, even those equipped with large language model (LLM) based text encoders, remain text–pixel mappers---they employ LLMs merely as text encoders, 
without leveraging their inherent reasoning capabilities to infer what should be visually depicted given the textual prompt. 
To move beyond such literal generation, we propose the \textbf{think-then-generate (T2G)} paradigm, where the LLM-based text encoder is encouraged to reason about and rewrite raw user prompts; the states of the rewritten prompts then serve as diffusion conditioning. 
To achieve this, we first activate the think-then-rewrite pattern of the LLM encoder with a lightweight supervised fine-tuning process.
Subsequently, the LLM encoder and diffusion backbone are co-optimized to ensure faithful reasoning about the context and accurate rendering of the semantics via \textbf{Dual-GRPO}. 
In particular, the text encoder is reinforced using image-grounded rewards to infer and recall world knowledge, while the diffusion backbone is pushed to produce semantically consistent and visually coherent images. 
Experiments show substantial improvements in factual consistency, semantic alignment, and visual realism across reasoning-based image generation and editing benchmarks, achieving 0.79 on WISE score, nearly on par with GPT-4o.
Our results constitute a promising step toward next-generation unified models with reasoning, expression, and demonstration capacities. 

\end{abstract}

%% file: sec/1_intro.tex
\section{Introduction}
\label{sec:intro}
Text-to-image (T2I) diffusion models (DMs) have demonstrated remarkable capabilities in generating high-fidelity and diverse images given textual prompts, as exemplified by Stable Diffusion~\cite{rombach2022high,esser2024scaling} and the FLUX series~\cite{flux-1,batifol2025flux,labs2025flux}.
To capture richer semantic representations of textual prompts, researchers have progressively adopted stronger text encoders—from early CLIP models~\cite{rombach2022high} to large language models (LLMs)~\cite{chenpixart,chen2024pixart} and their variants with visual inputs, i.e., vision language models (VLMs). 
Representative examples include the open-source Qwen-Image~\cite{wu2025qwen}, which employs Qwen2.5-VL~\cite{bai2025qwen2} as the encoder, and advanced commercial systems such as Seedream 4.0~\cite{seedream2025seedream}, GPT-4o~\cite{hurst2024gpt}, and Gemini-2.5-Flash-Image~\cite{google_gemini2_5_flash_image_2025}. 
The LLM encoder offers two key advantages: (1) native understanding of mixed text–image prompts, and (2) extensive world knowledge that enhances instruction following. 
These capabilities may alleviate the burden of iteratively crafting and refining prompts, as in prior work~\cite{omost,mo2024dynamic,xiang2025promptsculptor}, potentially enabling more natural interactions for visual generation.

\begin{figure}[t]
    \centering
    \makebox[\linewidth][c]{%
        \includegraphics[width=1.06\linewidth]{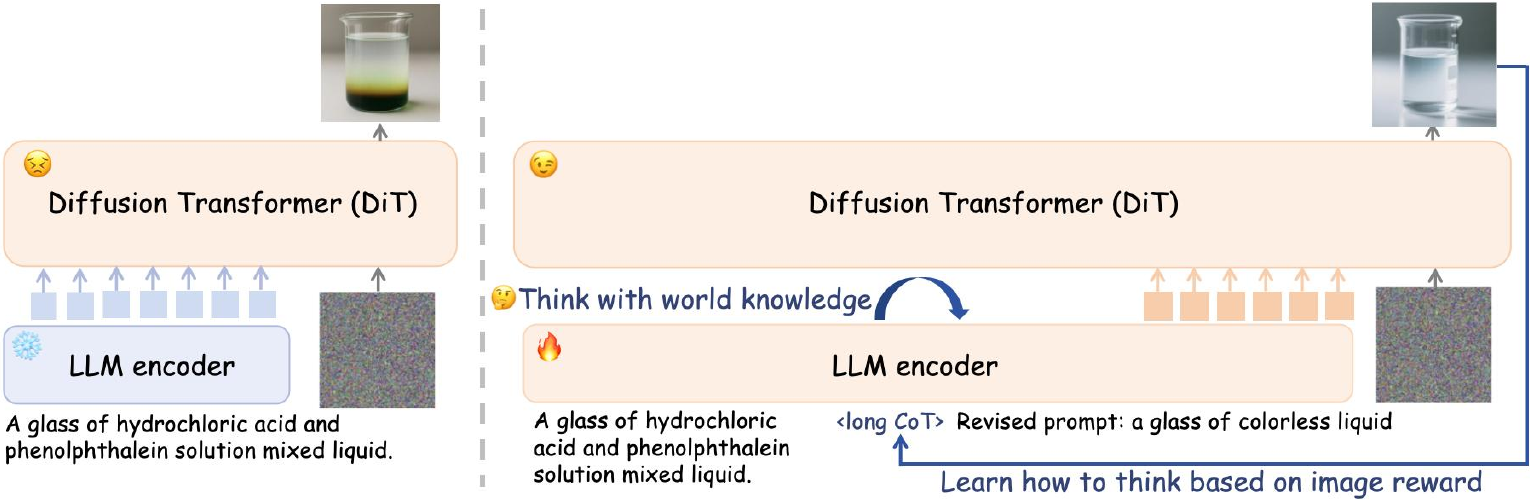}
    }
    \caption{An overview of our \textit{think-then-generate} method. Beyond using LLM as a frozen text encoder, we train it to think and refine the raw user prompts guided by the reward of output images.}
    \label{fig:pipe}
    \vspace{-0.6cm}
\end{figure}

\begin{figure*}
    \centering
    \includegraphics[width=1\linewidth]{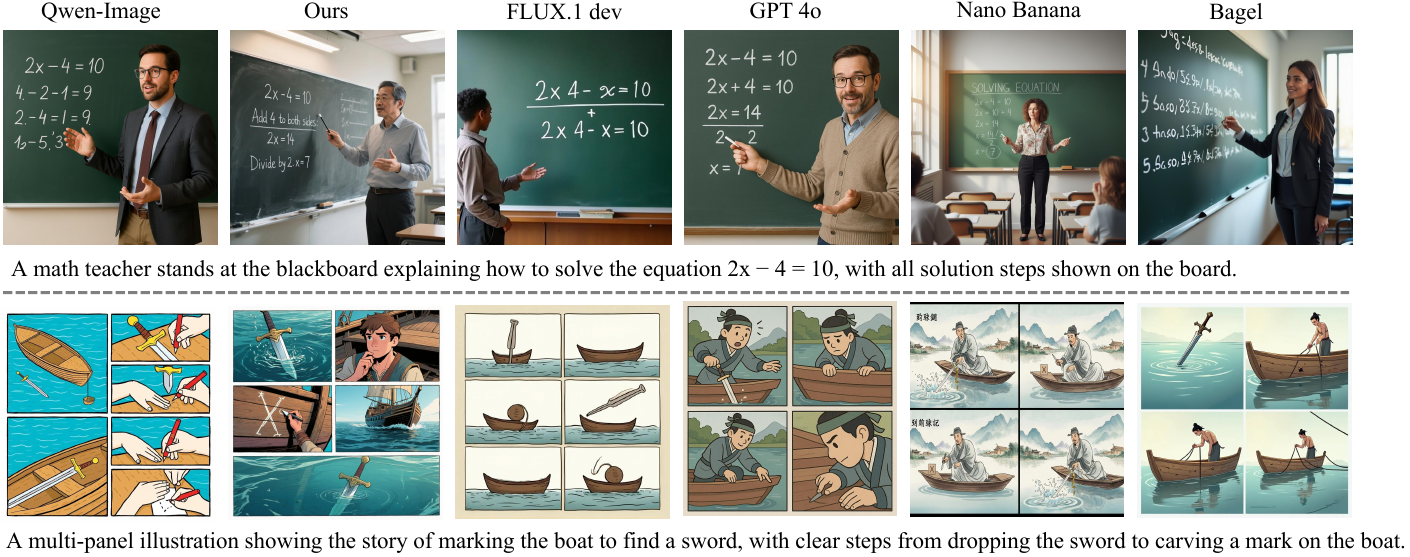}
    \caption{\textbf{Comparison of T2I models on conceptual visual generation.} Our Qwen-Image under \textit{think-then-generate} pipeline produces semantically aligned and visually coherent results, correctly interpreting user intent given prompt ``Holiday celebrating the birth of Jesus Christ.'' (\textit{e.g.}, generating a warm Christmas celebration rather than literally depicting Jesus), whereas vanilla Qwen-Image behaves like a simple text–pixel mapper and often fails to capture conceptual meanings.}
    \label{fig:demo_t2i}
\end{figure*}

However, in practice, existing models often fail to fully exploit the potential of LLMs for reasoning. 
Typically, the models are trained on large-scale descriptive image–caption pairs with the LLM encoder frozen~\cite{batifol2025flux,wu2025qwen,cai2025z} (i.e., solely as a feature extractor). 
Consequently, they can handle only literal and descriptive instructions (\textit{e.g.}, specific object colors or textures) but struggle with conceptual instructions (\textit{e.g.}, illustrating how a machine operates), as shown in Figure~\ref{fig:demo_t2i}.  
I.e., they function as simple \emph{text–pixel mappers}. 
Ideally, the LLM encoder should transcend prompt encoding to reason over raw instructions, leverage inherent world knowledge for conceptual tasks, and generate semantically enriched prompts for diffusion generation.
However, such a functionality cannot be trivially incentivized by current post-training frameworks for DMs~\cite{liu2025flow,zheng2025diffusionnft,yan2025unified,li2025branchgrpo}. 


This paper implements the \textit{think-then-generate (T2G)} paradigm for reasoning-aware text-to-image diffusion, where the LLM encoder is able to reason over and rewrite raw user prompts, with the embeddings of rewritten prompts fed to the diffusion transformer (DiT)~\cite{peebles2023scalable,esser2024scaling} as generation conditioning, as shown in Figure~\ref{fig:pipe}. 
To realize it, we first construct a supervised fine-tuning dataset that enriches raw user prompts with chain-of-thought (CoT) reasoning as well as rewritten prompts.
We finetune the LLM on it to acquire the think-then-rewrite pattern. 
We then co-optimize the LLM encoder and the DiT decoder via a \textit{Dual-GRPO} strategy, where the rewritten prompts act as the bridge between text reasoning and image synthesis and image-based rewards are maximized in an end-to-end manner. 
Considering the distinct roles of the encoder and the DiT for image generation, we tailored the reward objective to each component, where the former is optimized for \textit{semantic alignment and conceptual understanding} and the latter focuses on \textit{visual realism and aesthetic quality}. 
This optimization not only activates the world knowledge embedded in LLMs but also adapts DiT to the evolved representation space of the text encoder. 

Empirically, on the T2I task, we apply our method to the state-of-the-art open-source model Qwen-Image~\cite{wu2025qwen}, and observe a score of 0.79 on the WISE benchmark~\cite{niu2025wise}, which surpasses the pre-trained Qwen-Image by 30\%, and substantially outperforms other open-source models such as Bagel~\cite{deng2025emerging} and Janus-Pro-7B~\cite{chen2025janus}.
Notably, this performance is on par with the commercial GPT-4o~\cite{hurst2024gpt}. 
Moreover, our method achieves a score of 92.2 on T2I-ReasonBench~\cite{sun2025t2i}, surpassing the strong closed-source model Gemini-2.0~\cite{google_gemini2}.
For image editing, we apply our method to Qwen-Image-Edit~\cite{wu2025qwen}, and qualitative results demonstrate its ability to produce more faithful, fine-grained, and instruction-aligned modifications. 
Moreover, in more challenging and realistic scenarios—such as generating schematic illustrations of human activities (\textit{e.g.}, teaching math)—our trained model demonstrates superior knowledge grounding, visual plausibility, and aesthetic quality, highlighting its potential as next-generation unified models with reasoning, expression, and demonstration capacities in real-world applications.




%% file: sec/2_preliminary.tex
\section{Preliminary: Group Relative Policy \\Optimization}
\label{sec:preliminary}


Group Relative Policy Optimization (GRPO) is a reinforcement learning algorithm popularized in optimizing large generative models~\cite{shao2024deepseekmath, guo2025deepseek}. It can be viewed as a variant of policy gradient methods such as PPO~\cite{schulman2017proximal}, with introduced group-wise relative normalization of rewards to better handle high-variance signals during training.

Formally, given a group of rollout trajectories $\{\mathbf{o}_g\}_{g=1}^G$ sampled from the current policy $\pi_\theta$, unlike actor–critic methods, GRPO groups samples with similar input prompts or conditions, and then directly calculates relative advantages from normalized group rewards rather than learning them through a value model:
\begin{equation}
    \hat{A}_g = \frac{R_g - \operatorname{mean}\left( \{R_g\}_{g=1}^G \right)}{\operatorname{std}\left( \{R_g\}_{g=1}^G \right)},
\end{equation}
where $R_g$ is the scalar reward for the trajectory $\mathbf{o}_g$. This design eliminates the need for a value model, making GRPO more efficient and easier to scale for large models such as LLMs~\cite{havrilla2024teaching} and diffusion/flow matching models~\cite{liu2025flow,xue2025dancegrpo}. 
The typical GRPO objective is: 
\begin{align}
    & \mathcal{J}_{GRPO}(\theta) = \mathbb{E}_{\{\mathbf{o}_g\}_{g=1}^G \sim \pi_{\theta_{old}}} \\ \nonumber
    &\left[\frac{1}{G} \sum_{g=1}^G \min \left( r_g(\theta) \hat{A}_g, \text{clip}(r_g(\theta), 1-\epsilon, 1+\epsilon)\hat{A}_g \right)\right] \\ \nonumber
    &- \beta \mathbb{D}_{KL}\left[ \pi_\theta \parallel \pi_{\theta_{ref}} \right],
\end{align}
where $r_g(\theta) = \frac{\pi_\theta(\mathbf{o}_g)}{\pi_{\theta_{old}}(\mathbf{o}_g)}$, $\pi_{\theta_{ref}}$ is the reference policy, and $\beta$ is the KL divergence regularization parameter.

\textbf{GRPO for LLMs.}
For LLMs, the policy $\pi_\theta(o_t \mid o_{<t}, q)$ is a language model that generates text tokens $o_t$ conditioned on the previous tokens $o_{<t}$ and the user prompt $q$, this holds a tractable likelihood and is easy to compute. In typical GRPO applications, training relies on outcome-based rewards, where a reward model evaluates only the final generated text. All intermediate tokens in the rollout trajectory are assigned the same reward, which implicitly assumes that each token contributes equally to the final outcome.

\textbf{Flow-GRPO for Flow Matching Models.}
Although GRPO has been successfully applied to LLMs, applying it to flow matching models is nontrivial since their ODE-based sampling dynamics do not provide stochasticity to generate diverse trajectories for advantage estimation and policy exploration. 
Flow-GRPO~\cite{liu2025flow} addresses this limitation by transforming the deterministic Flow-ODE into an equivalent stochastic differential equation (SDE) and discretizing it via the Euler-Maruyama scheme. This yields the following transition kernel:
\begin{equation}
\pi_\theta(\mathbf{x}_{t-1} \mid \mathbf{x}_{t}) = \mathcal{N}(\mathbf{x}_{t-1}; \mu_\theta(\mathbf{x}_{t}), g_t^2 \Delta t \mathbf{I}),
\end{equation}
here $g_t = a \sqrt{\frac{t}{1 - t}}$ controls the level of stochasticity, $\Delta t$ is the discretization step size and $\mu_\theta(\mathbf{x}_{t})$ equals to:
\begin{equation}
    \mathbf{x}_t + \left[ \mathbf{v}_\theta(\mathbf{x}_{t}, t) + \frac{g_t^2}{2t}(\mathbf{x}_t + (1-t)\mathbf{v}_\theta(\mathbf{x}_{t}, t))\right]\Delta t.
\end{equation}
Crucially, the transition kernels are reduced to tractable Gaussian distributions, which enables the direct application of the GRPO policy update to flow matching models.

%% file: sec/3_method.tex
\section{Method}
\label{sec:method}
In this section, we first describe our data curation pipeline for constructing the supervised fine-tuning dataset used to train the LLM to replicate the \textit{T2G} pattern. We then introduce {Dual}-GRPO, a reinforcement learning strategy that jointly optimizes both the LLM and the DiT using image-based rewards.

\begin{figure}[t]
    \centering
    \includegraphics[width=0.8\linewidth]{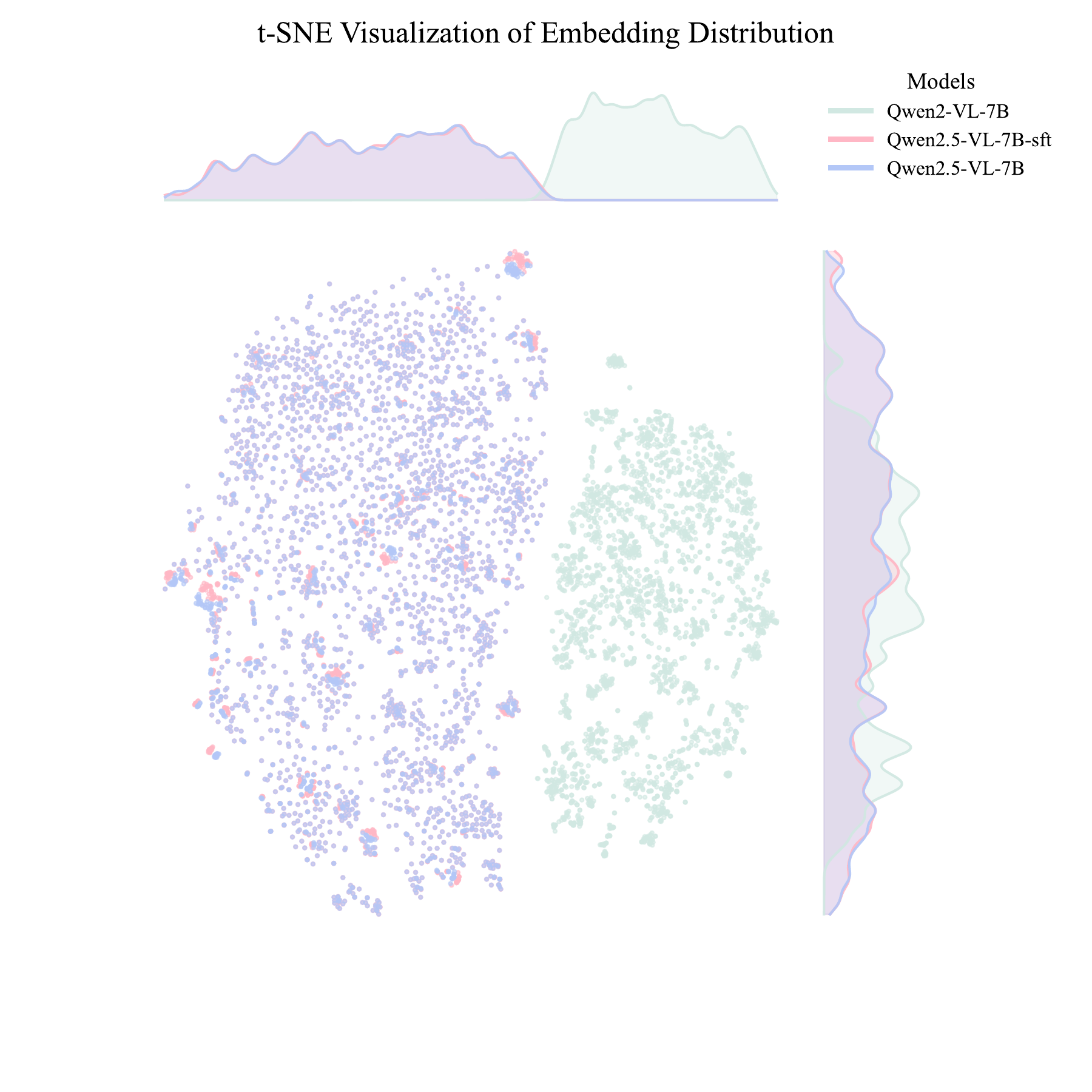}
    \caption{\textbf{t-SNE visualization of the embedding distributions before and after SFT.} The distributions remain highly consistent, indicating that our SFT procedure preserves the embedding space structure and thus maintains compatibility with the DiT, enabling it to render stable and coherent visual outputs.}
    \label{fig:tsne}
    \vspace{-0.5cm}
\end{figure}
\subsection{Reasoning-aware Behavior Activation}
Given a raw user prompt set $Q=\{q\}$, we aim to enable the model to follow a \textit{T2G} paradigm. Specifically, for each raw prompt, the model first performs CoT reasoning using its world knowledge to explicitly outline the content to be depicted, followed by summarizing this CoT into a descriptive refined prompt. During image generation, the embedding of the refined prompt serves as the conditioning signal for the DiT. To learn this pattern, we process $Q$ to construct a supervised fine-tuning dataset using Gemini-2.5~\cite{google_gemini2_5_flash_image_2025}. We prompt it to perform CoT reasoning, descriptively inferring what should be depicted and generating a refined prompt. This curated dataset follows the format: raw user prompt → [long CoT] → refined prompt. We then fine-tune the text encoder on this curated dataset.

Qwen2.5-VL serves as both a rewriter and an encoder in our \textit{T2G} paradigm. While SFT activates its rewriting potential, it remains unclear whether SFT degrades its encoding function: \textit{does SFT alter the embedding space in a way that harms the DiT’s ability to generate coherent images?} To examine this, we visualize Qwen2.5-VL~\cite{bai2025qwen2} text embeddings before and after SFT using t-SNE~\cite{cai2022theoretical}. It reveals distributional shifts by showing whether embeddings overlap or separate in a low-dimensional space. Interestingly, Figure~\ref{fig:tsne} shows that the embeddings totally overlap, indicating that SFT preserves the distribution of the embedding space and thereby maintains DiT’s capacity to produce stable and reasonable visual outputs.

\subsection{Dual-GRPO for LLM-DiT Composite Models}
\begin{figure*}[!htbp]
    \centering
    \includegraphics[width=1\linewidth]{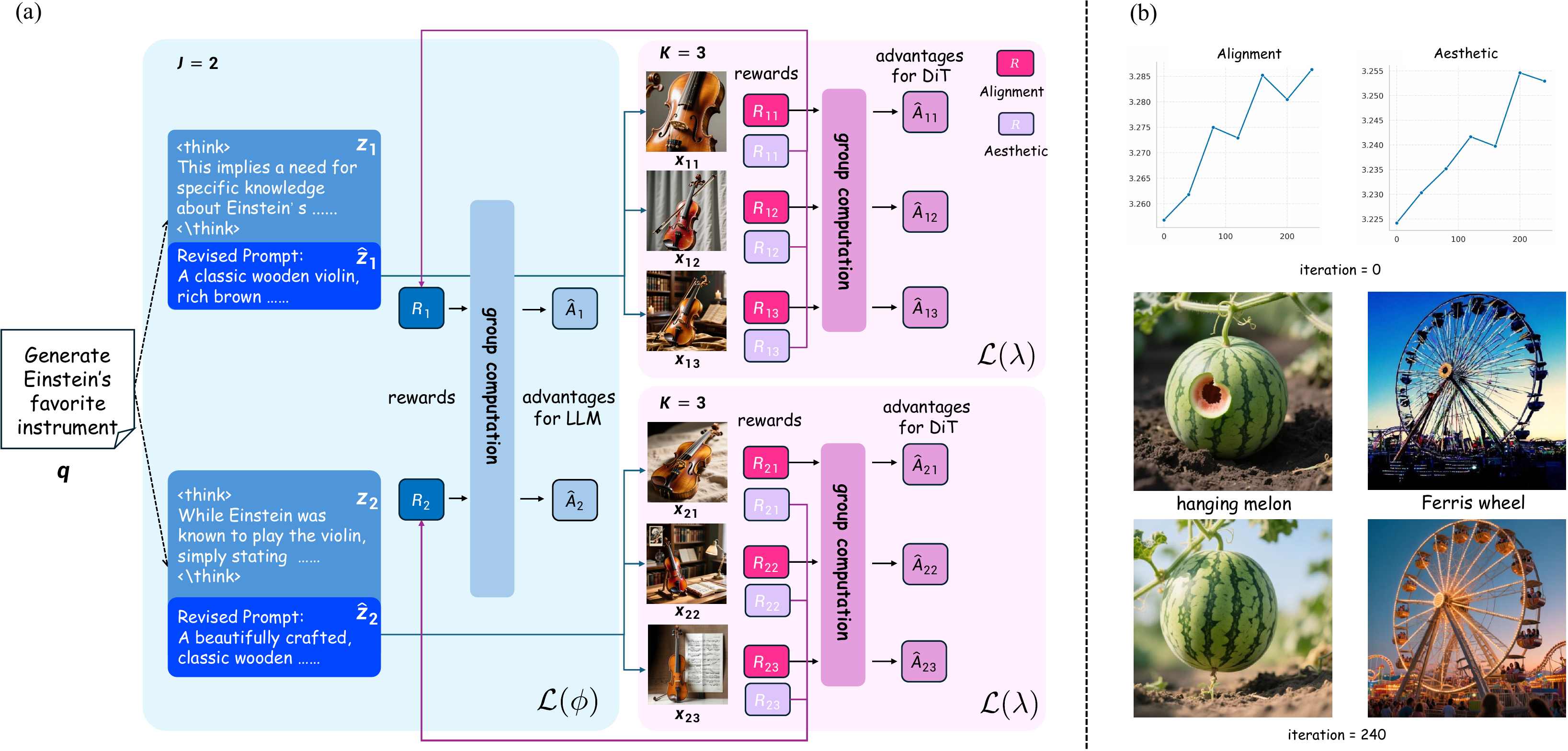}
    \caption{Dual-GRPO training trajectories. (a) Tree-structured rollout for a given user prompt $q$: the LLM encoder samples $J$ reasoning traces, each rewritten prompt conditions the DiT to generate $K$ images. Image-grounded rewards are aggregated to compute group-relative advantages for updating both the LLM and the DiT. (b) Evolution of alignment and style scores during training, demonstrating how DiT training improves both semantic alignment and visual quality over time. }
    \label{fig:tree_trajectory}
\end{figure*}
Given $q$, which may be vague or implicitly specified, conventional DMs employ an LLM text encoder $p_\phi$ to encode $q$ into a latent representation $p_\phi(q)$, which serves as the conditioning input for the DiT $p_\lambda$ to generate the corresponding image. After our supervised fine-tuning, the LLM encoder first performs CoT reasoning to generate text tokens $\mathbf{z}$ and then encodes them into a reasoning-aware representation $\hat{\mathbf{z}}$ that serves as conditions for the DiT decoding.

To be specific, the DMs can be viewed as a composite policy model parameterized by $\theta = \{\phi, \lambda\}$. Formally, it can be defined as
\begin{equation}
\pi_\theta(o_t \mid s_t) = 
\left\{
     \begin{array}{lr}
     p_\phi(z_t \mid z_{<t}, q)  &t \leq \ell, \\
     p_\lambda(x_{t} \mid x_{t-1}, \hat{\mathbf{z}}) &t > \ell.\\
     \end{array}
\right.
\end{equation}
And the rollout trajectory is represented as
$$
\mathbf{o} = \{z_1, \cdots, z_{\ell}, x_{\ell+1}, \cdots, x_{\ell+m}\}.
$$
In the above, $z_i$ denotes the $i$-th generated text token by LLM in $\mathbf{z}$, with reasoning budget $\ell$.
And $x_i$ refers to the $i$-th prediction during the reverse sampling process $\mathbf{x}$ with total iterative steps $m$.

To apply the policy gradient algorithm to the composite model $\pi_\theta$, we can optimize the following objective:
\begin{align}
    &\max_{\theta = \{\phi, \lambda\}} \mathbb{E}_{q\sim p(Q), \{z_1, \cdots, x_{\ell+m}\} \sim \pi_\theta} \\ \nonumber
    &\frac{1}{\ell+m} \left[ \sum_{t=1}^{\ell}  R_1(z_t, z_{<t}, q) + \sum_{t=\ell+1}^{\ell+m} R_2(x_t, x_{t-1}, \hat{\mathbf{z}}) \right],
\end{align}
where $p(Q)$ is the distribution of user prompts, $R_1$ and $R_2$ respectively denote the reward functions for the LLM and DiT components, details about the reward functions are discussed in Section~\ref{sec:reward}.

To take advantage of the group relative formulation for the estimation of advantages, we first sample a collection of outputs from the previous policy $\pi_{\theta_{old}}$. Specifically, as illustrated in Figure~\ref{fig:tree_trajectory}, given a user prompt $q$, we sample $J$ reasoning sequences $\{\mathbf{z}_j\}_{j=1}^J$ from $p_{\phi_{old}}$, and for each $\hat{\mathbf{z}}_j$, we sample $K$ images $\{\mathbf{x}_{j,k}\}_{k=1}^K$ from $p_{\lambda_{old}}$. This hierarchical sampling strategy enables us to compute group-relative advantages for both components within one rollout. By incorporating the standard PPO clipping mechanism and KL divergence regularization, we arrive at the Dual-GRPO objective:
\begin{align}
    &\max_{\theta = \{\phi, \lambda\}} \mathbb{E}_{q\sim p(Q), \{z_1, \cdots, x_{\ell+m}\}_{i=1}^{J \times K} \sim \pi_{\theta_{old}}} \\ \nonumber
    &\frac{1}{l+m} \left[ \sum_{t=1}^{l} \mathcal{L}_t(\phi) + \sum_{t=l+1}^{l+m} \mathcal{L}_t(\lambda) \right],
\end{align}
where $\mathcal{L}_t(\phi)$ follows the standard GRPO formulation for language models with group size $J$:
\begin{align}
    &\frac{1}{J}\sum_{j=1}^J
    \left[ \min \left( r_{j,t}(\phi), \text{clip}(r_{j,t}(\phi), 1-\epsilon, 1+\epsilon) \right) \hat{A}_{j, t} \right] \\\nonumber
    &- \beta \mathbb{D}_{KL}\left[ p_\phi(z_{j, t}) \parallel p_{\phi_{ref}}(z_{j, t}) \right],
\end{align}
where $r_{j,t}(\phi) = \frac{p_\phi(z_{j,t} \mid z_{j,<t}, q)}{p_{\phi_{old}}(z_{j,t} \mid z_{j,<t}, q)}$, the group-relative advantage $\hat{A}_{j, t}$ for the LLM $p_\phi$ is calculated by aggregating the rewards from all $K$ images generated from the same reasoning output $\hat{\mathbf{z}}_j$.

Similarly, denote $r_{j,k,t}(\lambda) = \frac{p_{\lambda}(x_{j,k,t} \mid x_{j,k,t-1}, \hat{\mathbf{z}}_j)}{p_{\lambda_{old}}(x_{j,k,t} \mid x_{j,k,t-1}, \hat{\mathbf{z}}_j)}$, the loss term $\mathcal{L}_t(\lambda)$ for the DiT $p_\lambda$ employs a formulation of Flow-GRPO with an extra batch dimension with size $J$ and group size $K$:
{\small
\begin{align}
    &\frac{1}{J}\sum_{j=1}^J \frac{1}{K} \sum_{k=1}^K \left[ \min \left( r_{j,k,t}(\lambda), \text{clip}(r_{j,k,t}(\lambda), 1-\epsilon, 1+\epsilon) \right) \hat{A}_{j, k, t} \right] \\\nonumber
    &- \beta \mathbb{D}_{KL}\left[ p_\lambda(x_{j, k, t} \mid x_{j, k, t-1}, \hat{\mathbf{z}}_j) \parallel p_{\lambda_{ref}}(x_{j, k, t} \mid x_{j, k, t-1}, \hat{\mathbf{z}}_j) \right].
\end{align}
}

\subsection{The Reward Function and Scheduler}
\label{sec:reward}
When GRPO is applied to language models or flow matching models, it is common to use an outcome-based reward that depends only on the final prediction, and all intermediate states within the rollout trajectory share the same reward. Such a design implicitly assumes that each step contributes equally to the final outcome, overlooking the fact that different stages of the trajectory influence distinct aspects of the final result.

Taking into account that the composite model is a two-stage model, we can design stage-specific reward functions that align with the distinct characteristics of each stage.

During the LLM reasoning stage, most of the semantic content of the generated image is determined within the reasoning output $\hat{\mathbf{z}}$. And the reward for each reasoning step is the averaged semantic consistency score $R_{sem}$ over all $K$ generated images sampled from $p_{\lambda_{old}}(\mathbf{x} \mid \hat{\mathbf{z}})$:
\begin{equation}
    R_1(z_{j,t}, z_{j,<t}, q) = \beta_1(\tau)\frac{1}{K} \sum_{k=1}^K R_{sem}(\mathbf{x}_{j,k} , q).
\end{equation}
In the above, $\beta_1(\tau)$ is the scheduler for reward weighting, which is a function of the current training step $\tau$. With this scheduler, we can assign different emphases to the rewards of the LLM and the diffusion model at different stages of training. The corresponding advantage is calculated as
\begin{equation}
    \hat{A}_{j, t} = \frac{R_1(z_{j,t}, z_{j,<t}, q) - \operatorname{mean}\left( \{R_1(z_{j,t}, z_{j,<t}, q)\}_{j=1}^J \right)}{\operatorname{std}\left( \{R_1(z_{j,t}, z_{j,<t}, q)\}_{j=1}^J \right)}.
\end{equation}

During the diffusion sampling stage, the model needs to render the reasoning output into an aesthetically pleasing and physically consistent image. For a reverse time trajectory $\mathbf{x}_{j, k}$ conditioned on a previous reasoning output $\hat{\mathbf{z}}_j$, the reward for each step is defined as a weighted sum of the aesthetic score $R_{aes}$, the physical consistency score $R_{con}$, and the semantic consistency score $R_{sem}$ of the final generated image~\cite{sun2025t2i}:
\begin{align}
    &R_2(x_{j,k,t}, x_{j,k,t-1}, \hat{\mathbf{z}}_j) = \\ \nonumber
    &\beta_2(\tau) \left( \omega_1 R_{aes}(\mathbf{x}_{j,k}) + \omega_2 R_{con}(\mathbf{x}_{j,k}) + \omega_3 R_{sem}(\mathbf{x}_{j,k}) \right),
\end{align}
$\omega_1, \omega_2, \omega_3$ are the weights for different components of the reward. Similarly, $\beta_2(\tau)$ is the scheduler for the diffusion sampling stage.

%% file: sec/4_experiment.tex
\section{Experiments}
\label{sec:experiment}
In this section, we evaluate the effectiveness of our \textit{T2G} paradigm across T2I and image editing tasks. Both quantitative and qualitative results validate the advantages of our approach.
\subsection{Implementation Details}
We post-train Qwen-Image~\cite{wu2025qwen} and Qwen-Image-edit~\cite{wu2025qwen} for the T2I and image-editing tasks, respectively. Consequently, the LLM encoder is initialized from Qwen2.5-VL~\cite{bai2025qwen2}, and the corresponding DiT backbone is initialized from MM-DiT~\cite{esser2024scaling}. Our post-training pipeline is separated into two stages. For the supervised fine-tuning stage, each model is trained with a learning rate of 5e-6 and a batch size of 32. In the subsequent \textit{Dual}-GRPO stage, we apply a simple reward scheduler with constant and balanced weights: $\beta_1(\tau)=\beta_2(\tau)=0.5$, where both the LLM and the DiT are updated jointly at every iteration. In Appendix~\ref{append:exp_schedule}, we explore different designs of the scheduler. During the LLM training, we employ a learning rate of 2e-6 with a batch size of 256, generating 5 responses per input. For the DiT training, we adhere to the default configuration of FlowGRPO-fast\cite{liu2025flow}, utilizing a learning rate of 3e-4, a batch size of 32, and a clipping range of 1e-4. The generation process involves sampling 16 images for each prompt over 10 inference steps, with an SDE window of 2. 
\begin{table*}[]
\centering
\caption{\textbf{Comparison of image generation models on WISE.} Numbers in bold indicate the highest score among open-source models. Underlined numbers denote the highest score.}
\begin{tabular}{llccccccc}
\toprule
\multicolumn{1}{l}{Model} & Type & Cultural & Time & Space & Biology & Physics & Chemistry & \textbf{Overall $\uparrow$} \\ \midrule
FLUX.1-dev & \textit{diffusion}               & 0.48     & 0.58 & 0.62  & 0.42    & 0.51    & 0.35      & 0.50                        \\
SD-3.5-medium & \textit{diffusion}            & 0.43     & 0.50 & 0.52  & 0.41    & 0.53    & 0.33      & 0.45                        \\
SD-3.5-large & \textit{diffusion}             & 0.44     & 0.50 & 0.58  & 0.44    & 0.52    & 0.31      & 0.46                        \\
Bagel w/CoT  & \textit{unified}             & 0.76     & 0.69 & 0.75  & 0.65    & 0.75    & 0.58      & 0.70                        \\
Janus-Pro-7B  & \textit{unified}            & 0.30     & 0.37 & 0.49  & 0.36    & 0.42    & 0.26      & 0.35                        \\
HunyuanImage-3.0  & \textit{unified}   & 0.58     & 0.57 & 0.72  & 0.56    & 0.68    & 0.35      & 0.58               \\ 
T2I-R1 & \textit{unified}            & 0.56     & 0.55 & 0.63  & 0.54    & 0.55    & 0.30      & 0.54                        \\
Uni-CoT     & \textit{unified}              & 0.76     & 0.70 & 0.76  & 0.73    & 0.81    & 0.73      & 0.75                        \\
GPT 4o      & \textit{proprietary}              & \underline{0.81}     & 0.71 & \underline{0.89}  & \underline{0.83}    & 0.79    & \underline{0.74}      & \underline{0.80}                        \\
Qwen-Image     & \textit{diffusion}           & 0.62     & 0.63 & 0.78  & 0.55    & 0.67    & 0.35      & 0.61                        \\
\rowcolor[HTML]{ECF4FF} 
Ours (w/o SFT+GRPO)  & \textit{diffusion}   & 0.68     & 0.58 & 0.77  & 0.62    & 0.76    & 0.41      & 0.65                        \\
\rowcolor[HTML]{ECF4FF} 
Ours (w/o GRPO)   & \textit{diffusion}      & 0.76     & 0.66 & 0.79  & 0.74    & 0.84    & 0.65      & 0.74                        \\
\rowcolor[HTML]{ECF4FF} 
Ours  & \textit{diffusion}   & \textbf{0.80}     & \underline{\textbf{0.74}} & \textbf{0.83}  & \textbf{0.81}    & \underline{\textbf{0.85}}    & \textbf{0.66}      & \textbf{0.79}                        \\
\bottomrule
\end{tabular}
\label{tab:wise}
\end{table*}


\subsection{Text-to-Image Generation}
\subsubsection{Supervised Fine-tuning}
To activate the \textit{T2G} pattern of Qwen2.5-VL, we first carefully curate a raw user prompt dataset comprising 7,000 samples. A core criterion for curation is that each prompt requires integration of world knowledge and reasoning to generate a semantically coherent corresponding image. For example, prompts like ``Generate an image of traditional food associated with the Dragon Boat Festival''. Then we instruct Gemini-2.5~\cite{google_gemini2_5_flash_image_2025} to perform CoT reasoning using its world knowledge to explicitly outline what content should be depicted and summarize it into a descriptive refined prompt. We then fine-tune Qwen2.5-VL on this dataset to replicate the \textit{T2G} pattern when given new prompts.


\subsubsection{Quantitative Results}
\textbf{Benchmarks.} We evaluate performance across various T2I benchmarks requiring reasoning capacity. WISE~\cite{niu2025wise} evaluates models with 1000 meticulously crafted prompts across 25 sub-domains in cultural common sense, spatio-temporal understanding, and natural science. T2I-ReasonBench~\cite{sun2025t2i} comprises 800 prompts organized into four dimensions: idiom interpretation, textual image design, entity-reasoning, and scientific-reasoning. It introduces a two-stage evaluation framework to quantify performance: an LLM first generates prompt-specific question-criterion pairs evaluating if the image includes the essential elements resulting from correct reasoning, and a multimodal LLM then scores the generated image against these criteria.

\textbf{Baselines.} We compare with 10 state-of-the-art T2I models, including 4 diffusion-based
T2I models, 4 unified multimodal models, and 2 proprietary models. The diffusion-based T2I models are FLUX.1-dev~\cite{flux-1}, Stable-Diffusion-3-Medium~\cite{esser2024scaling}, Stable-Diffusion-3.5-Large~\cite{esser2024scaling}, and Qwen-Image~\cite{wu2025qwen}. The unified multimodal models are Bagel~\cite{deng2025emerging}, Uni-CoT~\cite{qin2025uni}, Emu3~\cite{wang2024emu3}, Janus-Pro-7B~\cite{chen2025janus}, HunyuanImage-3.0~\cite{cao2025hunyuanimage}, and T2I-R1~\cite{jiang2025t2i}. The proprietary models include Gemini-2.0~\cite{google_gemini2} and GPT-4o~\cite{hurst2024gpt}.

\begin{figure*}
    \centering
    \includegraphics[width=1\linewidth]{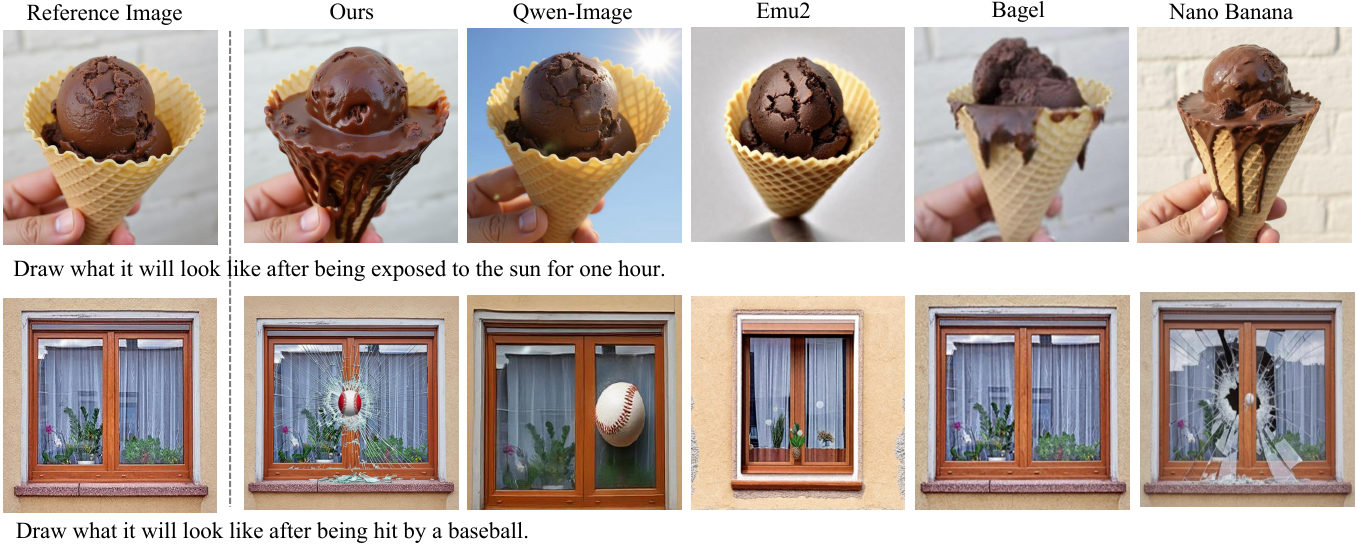}
    \caption{\textbf{Comparison of T2I models on conceptual image editing.} Vanilla Qwen-Image fails to interpret instructions (e.g., showing an ice cream under sunlight instead of melting), behaving as a text–pixel mapper. Our model correctly infers intended semantics, producing coherent, aesthetically pleasing edits with high consistency to the original image, outperforming unified models like Emu2 and Bagel.}
    \label{fig:demo_editing}
\end{figure*}

\begin{table*}[]
\centering
\caption{\textbf{Comparison of generation models on T2I-ReasonBench.} Numbers in bold are the highest score among open-source models. Underlined numbers denote the highest score.}
\begin{tabular}{llcccccccccc}
\toprule
\multicolumn{1}{c}{}      & \multicolumn{1}{c}{}                  & \multicolumn{2}{c}{Idiom} & \multicolumn{2}{c}{Textual} & \multicolumn{2}{c}{Entity} & \multicolumn{2}{c}{Scientific} & \multicolumn{2}{c}{\textbf{Overall $\uparrow$}} \\
\multicolumn{1}{l}{\multirow{-2}{*}{Model}} & \multicolumn{1}{l}{\multirow{-2}{*}{Type}} & Acc.        & Qual.       & Acc.         & Qual.        & Acc.         & Qual.       & Acc.          & Qual.          & Acc.                   & Qual.                  \\ \midrule
FLUX.1-dev  & \textit{diffusion}                                & 39.1        & 83.4        & 56.9         & 76.5         & 45.1         & 90.6        & 46.7          & 80.9           & 47.0                   & 82.8                   \\
SD-3.5-medium    & \textit{diffusion}                           & 34.4        & 80.6        & 58.0         & 70.1         & 44.8         & 92.1        & 49.9          & 83.0           & 46.8                   & 81.4                   \\
SD-3.5-large     & \textit{diffusion}                           & 35.6        & 85.3        & 62.2         & 75.4         & 46.6         & 92.6        & 52.9          & 84.5           & 49.3                   & 84.4                   \\
Bagel w/CoT       & \textit{unified}                          & 44.6        & 84.3        & 44.0         & 73.7         & 52.4         & 91.6        & 57.7          & 88.3           & 49.7                   & 84.5                   \\
Janus-Pro-7B    & \textit{unified}                            & 25.5        & 78.0        & 37.2         & 70.9         & 38.5         & 87.6        & 44.9          & 77.8           & 36.5                   & 78.6                   \\
    HunyuanImage-3.0       & \textit{unified}                    & 25.4        & 80.2        & 54.2         & 80.9         & 52.3        & 92.2         & 56.8          & 84.4           & 47.2         & 84.4         \\
UniCoT & \textit{unified}                            & 49.0       & 84.2        & 58.1         & 92.3         & 73.5         & 92.9       & 51.9         & 71.7           & 58.1                   & 85.3                   \\
Gemini-2.0      & \textit{proprietary}                            & 52.4        & 87.8        & 73.0         & 83.3         & 67.0         & 94.3       & 66.7          & 89.3          & 64.8                   & 88.7                   \\
GPT-4o          & \textit{proprietary}                            & \underline{75.7}        & \underline{94.5}        & \underline{86.9}         & \underline{97.6}         & \underline{77.5}         & \underline{96.6}        & \underline{74.7}          & \underline{94.3}           & \underline{78.7}                   & \underline{95.8}           \\
Qwen-Image        & \textit{diffusion}                           & 48.1        & 84.3        & 66.5         & 85.8         & 57.1         & 84.7        & 59.5          & 85.3           & 57.8                   & 87.5                   \\
\rowcolor[HTML]{ECF4FF} 
Ours (w/o SFT+GRPO)   & \textit{diffusion}                     & 51.7        & 86.3        & 71.7         & 83.8         & 57.3         & 92.8        & 62.8          & 87.6           & 60.9                   & 86.6                   \\
\rowcolor[HTML]{ECF4FF} 
Ours (w/o GRPO)    & \textit{diffusion}                         & 59.5        & 90.4        & 71.5         & 87.1         & 61.29        & 93.9        & 72.2          & 91.8           & 66.1                   & 90.8                   \\
\rowcolor[HTML]{ECF4FF} 
Ours     & \textit{diffusion}                   & \textbf{58.5}        & \textbf{90.6}        & \textbf{75.2}         & \textbf{89.5}         & \textbf{68.8}         & \textbf{95.2}        & \textbf{72.9}          & \textbf{93.5}           & \textbf{68.3}                   & \textbf{92.2}               \\
\bottomrule
\end{tabular}
\label{tab:t2i_reason}
\end{table*}

\textbf{Results analysis.} From Table~\ref{tab:wise} and \ref{tab:t2i_reason}, we find that DMs exhibit weak performance on both the WISE and T2I-ReasonBench benchmarks. This stems from its fundamental misalignment between their architectural design (i.e., treat LLMs only as a text encoder) and the demands of knowledge-intensive, reasoning-driven T2I generation. These models excel at pixel-level detail synthesis, but their reliance on shallow text-image alignment restricts them to capturing surface-level semantic correlations—insufficient for tasks requiring deep knowledge integration. For instance, DMs such as vanilla Qwen-Image only achieves 0.35 in WISE chemistry domains, failing to translate abstract concepts (e.g., reaction stoichiometry) into logically consistent visuals, underscoring the importance of our work.

Moreover, we evaluate the zero-shot performance of Qwen-Image by introducing a CoT step: for raw user prompts, we first prompt the model to generate reasoning before outputting a refined prompt. As shown in the results, this modification yields marginal improvement: the overall WISE score increases from 0.61 to 0.65. However, it still lags significantly behind optimized variants, as it lacks task-specific supervised signals to align its reasoning process with the need to infer visually actionable attributes suitable for DiT decoding.

To address this misalignment between reasoning and visual generation requirements, we further conduct SFT training. However, the CoT generation process remains unaware of the image generation module DiT under the SFT pipeline. This decoupling can lead to the LLM generating some strange tokens that the DiT cannot render into a reasonable image. This can be alleviated by further dual-GRPO training, where the image-grounded rewards are used for LLM optimization and DiT is adapted for better visual rendering. As seen, our models after SFT and GRPO achieves breakthrough performance on both benchmarks (0.79 on WISE, 68.3 accuracy on T2I-ReasonBench) by addressing the long-standing challenge of reasoning-generation co-optimization. It achieves a leading performance across open-source T2I models and outperforms the leading proprietary model Gemini-2.0, approaching the SOTA of closed systems GPT-4o, proving that with our \textit{T2G} paradigm, open models can match or even surpass closed alternatives in knowledge-intensive visual generation tasks.

\subsubsection{Qualitative Results}
We analyze the CoT trajectories generated before and after GRPO and observe errors in the roll-out samples (e.g., labeling dumplings as a traditional food for the Dragon Boat Festival in China). These incorrect samples are penalized during GRPO, leading to improved accuracy. Moreover, Figure~\ref{fig:demo_t2i} illustrates images generated by Qwen-Image under the \textit{T2G} paradigm, demonstrating its superior ability to infer and understand raw user prompts, while producing text-consistent, visually coherent, and aesthetically pleasing images.

\subsection{Image Editing}
\subsubsection{Supervised Fine-tuning}
For raw user instructions for image editing, we select the UniREdit-Data-100K~\cite{qin2025uni}, a large-scale synthetic dataset specifically curated for image editing with high-quality CoT reasoning annotations. Following the same training data format as T2I generation tasks, we augment this dataset by instructing Gemini-2.5~\cite{google_gemini2_5_flash_image_2025} to conclude a refined prompt for each raw instruction and corresponding CoT. This augmented dataset is used to fine-tune Qwen-Image-Edit. For \textit{Dual}-GRPO training, we randomly select 5k raw user prompts from this dataset. 

\subsubsection{Quantitative Results}
\textbf{Benchmarks.} We evaluate performance across various image editing benchmarks requiring reasoning capacity. UniREditBench~\cite{qin2025uni} evaluates models with 2,700 meticulously curated samples, covering both real- and game-world scenarios. RISEBench~\cite{zhao2025envisioning} comprises 327 prompts designed to evaluate models on reasoning-aware tasks across temporal, causal, spatial, and logical dimensions. It proposes a robust evaluation framework that assesses instruction reasoning, appearance consistency, and visual plausibility using the LMM-as-a-judge approach.

\textbf{Results analysis.} As shown in Table~\ref{tab:edit}, after our post-training, QwenImage-Edit achieves strong performance on image editing tasks, even surpassing Gemini-2.5-Flash-Image on the UniREdit benchmark. Moreover, it exhibits a substantial improvement over the SFT-only baseline, demonstrating the effectiveness of our \textit{Dual}-GRPO in enhancing CoT reasoning through image-based rewards.

\begin{table}[t]
    \centering
    \caption{\textbf{Comparisons of image editing results on UniREditBench and RISEBench.} Numbers in bold indicate the highest score among open-source models. Underlined numbers denote the highest score. Detailed scores are shown in Appendix~\ref{append:detailed_edit}.}
    \resizebox{0.47\textwidth}{!}{
    \begin{tabular}{lccc}
    \toprule
    \multicolumn{1}{c}{ Model} &  Type                                                  & UniREdit $\uparrow$  & RISE $\uparrow$ \\ \midrule
    Gemini-2.5-Flash-Image    &  \textit{proprietary}                                                     & 68.3     & \underline{32.8}                 \\
    GPT-Image-1               &   \textit{proprietary}                                                    & \underline{73.4}     & 28.9                \\
    Seedream-4.0              & \textit{proprietary}                         & 55.8     &   10.8         \\
    Bagel w/ CoT            &           \textit{unified}                    & 51.0      & 9.2         \\
    UniWorld-V2 &  \textit{unified} & 54.9 & - \\
    FLUX.1-Kontext        & \textit{diffusion} & 45.8      & 5.8     \\ 
    Qwen-Image-Edit           &        \textit{diffusion}                      & 56.5      & 8.9          \\
    \rowcolor[HTML]{ECF4FF} 
    Ours (w/o GRPO)                      & \textit{diffusion}                 &    61.1  &      20.2                        \\
    \rowcolor[HTML]{ECF4FF} 
    Ours                      & \cellcolor[HTML]{ECF4FF}   \textit{diffusion}                 &    \textbf{68.7}  &      \textbf{23.9}                     \\
    \bottomrule
    \end{tabular}
    }
    \label{tab:edit}
\end{table}

\subsubsection{Qualitative Results}
As shown in Table~\ref{fig:demo_editing}, vanilla Qwen-Image struggles to interpret conceptual editing instructions. For the prompt ``draw what the ice cream looks like after being exposed to the sun,'' it merely renders the reference image under sunlight, resulting in visually incoherent output and revealing its behavior as a simple text–pixel mapper. In contrast, our model trained under the \textit{T2G} paradigm correctly infers the intended semantics—\textit{e.g.}, depicting the ice cream melting—and produces coherent edits. Moreover, the qualitative results show that our model generates more aesthetically pleasing images than unified models such as Emu2 and Bagel, while achieving the highest consistency between the edited image and the original one. These findings highlight its potential as a next-generation unified model for real-world applications.

\subsection{Ablation Studies}
To evaluate the effectiveness of our SFT, we conduct an ablation study, noting that the vanilla Qwen-VL already possesses a certain degree of reasoning capability. Specifically, we compare models trained with and without the SFT stage that teaches the \textit{T2G} pattern. As shown in Table~\ref{tab:ablation}, the model with SFT significantly outperforms the variant without it. This improvement stems from SFT explicitly aligning the reasoning process (CoT → refined prompt) with the generation pipeline, enabling the model to produce semantically grounded refined prompts rather than relying on shallow pattern matching or implicit reasoning. Consequently, the DiT receives more coherent conditioning signals, leading to consistently higher-quality image generation.

\begin{table}[t]
    \centering
    \caption{Ablation studies on the importance of SFT. w/o SFT denotes the vanilla Qwen-Image with thinking system prompt.}
    \begin{tabular}{lccc}
        \toprule
        Model & WISE$\uparrow$ & \multicolumn{2}{c}{T2I-Reason$\uparrow$} \\
        \cmidrule(r){3-4}
        & & Acc. & Qual. \\
        \midrule
        w/o SFT       & 0.65 & 60.9 & 86.6 \\
        w/o SFT + GRPO  & 0.70 & 66.9 & 91.0 \\
        w/ SFT             & 0.74 & 66.1 & 90.8 \\
        w/ SFT + GRPO        & \textbf{0.79} & \textbf{68.3} & \textbf{92.2} \\
        \bottomrule
    \end{tabular}
    \vspace{-4pt}
    \label{tab:ablation}
\end{table}



%% file: sec/5_related.tex
\section{Related Work}
\label{sec:related}

\subsection{RL for Diffusion Models}
Reinforcement learning has emerged as a powerful paradigm for aligning generative models with human preferences, yet its application to diffusion models faces unique challenges due to deterministic sampling and intractable likelihood estimation. Early approaches like DDPO~\cite{black2023training} discretized the reverse process to enable policy gradient updates but suffered from solver restrictions and instability when scaling to complex prompts. Flow-GRPO~\cite{liu2025flow} pioneered online RL integration for flow matching by converting ODEs to equivalent SDEs for exploration and introducing denoising reduction to improve sampling efficiency. While effective for compositional generation and text rendering, it optimizes only the diffusion backbone while freezing the text encoder—a limitation restricting its capacity for knowledge-intensive tasks. Similarly, DanceGRPO~\cite{xue2025dancegrpo} enhanced stability through group-wise relative policy optimization across multiple generative paradigms but maintained the same encoder-freezing paradigm. DiffusionNFT~\cite{zheng2025diffusionnft} circumvented likelihood estimation by optimizing directly on the forward process via contrastive preference learning, achieving remarkable efficiency gains. However, all these methods treat the text encoder as a static feature extractor, overlooking its potential for semantic reasoning when endowed with world knowledge (e.g., in LLM-based encoders). Recent work like Diffusion-SDPO~\cite{fu2025diffusion} addressed optimization pathologies in preference learning but still operated solely on the decoder. Our \textit{Dual}-GRPO framework fundamentally diverges by jointly optimizing both the LLM encoder and diffusion decoder through a composite policy design.

\subsection{Multimodal Models for Image Generation}
The shift toward unified multimodal architectures has accelerated with the emergence of models possessing strong image generation capabilities, such as HunyuanImage~\cite{cao2025hunyuanimage} and BAGEL~\cite{deng2025emerging}, which integrate vision-language understanding and generation within a single autoregressive framework. HunyuanImage leverages MoE architectures (over 80B parameters) to unify multimodal processing, while BAGEL demonstrates emergent reasoning capabilities via training on interleaved text-image-video data. This deep integration of textual and visual modeling holds promise for improved image generation conditioned on conceptual and complex prompts. Despite these advances, such models remain biased toward literal generation, largely due to their reliance on descriptive caption datasets during training. To mitigate this, several post-training frameworks incorporate multimodal CoT reasoning. For example, T2I-R1~\cite{jiang2025t2i} leverages reinforcement learning with bi-level CoT reasoning to jointly enhance high-level prompt planning and low-level pixel generation. In parallel, Uni-CoT proposes a unified two-level reasoning paradigm, consisting of a macro-Level CoT for high-level task planning and a micro-Level CoT for fine-grained subtask execution, all within a single multimodal model. Instead of focusing on unified architectures, our framework emphasizes incorporating text-based CoT reasoning within the LLM encoder to enhance diffusion models.


%% file: sec/6_conclusion.tex
\section{Conclusion}
\label{sec:conclusion}
This work addresses a key limitation of T2I diffusion models: their literal text-to-pixel mapping fails to exploit LLMs’ reasoning and world knowledge. We propose the \textit{think-then-generate} paradigm, turning LLMs from passive encoders into active reasoning agents. Our framework has two components: a lightweight supervised fine-tuning to activate LLMs reasoning with raw prompts; and \textit{Dual}-GRPO, which jointly optimizes LLMs for image-grounded semantic consistency and DiT for visual realism. Experiments show that our approach outperforms open-source baselines, surpasses Gemini-2.0, and excels in conceptual image editing, enabling knowledge-intensive visual generation. 





%% file: sec/X_suppl.tex
\clearpage
\setcounter{page}{1}
\setcounter{section}{0}  
\maketitlesupplementary
\begin{table}[t]
\centering
\caption{Comparison of different reward schedulers on T2I-ReasonBench.}
\begin{tabular}{l cc cc}
\toprule
& \multicolumn{2}{c}{Balanced Scheduler}
& \multicolumn{2}{c}{Staged Scheduler} \\
\cmidrule(lr){2-3} \cmidrule(lr){4-5}
Category & Acc. & Qual. & Acc. & Qual. \\
\midrule
Idiom        & 62.1 & 90.6 & 58.5 & 90.6 \\
Textual      & 77.5 & 90.1 & 75.2 & 89.5 \\
Entity       & 67.1 & 96.0 & 68.8 & 95.2 \\
Scientific   & 72.4 & 93.6 & 72.9 & 93.5 \\
\midrule
Overall $\uparrow$ & 69.8 & 92.6 & 68.3 & 92.2 \\
\bottomrule
\end{tabular}
\label{tab:schedule_ablate}
\end{table}


\section{Ablation on Different Reward Schedulers}
\label{append:exp_schedule}
In Section 3.3, we introduce the reward-weighting scheduler used during training. Specifically, we adopt a balanced scheduler with constant weights, setting $\beta_1(\tau) = \beta_2(\tau) = 0.5$, such that both the LLM and the DiT are updated jointly at every iteration. In this section, we an alternative staged piecewise scheduler, where we set $\beta_1(\tau)=1, \beta_2(\tau)=0$ at early training steps and switch to $\beta_1(\tau)=0, \beta_2(\tau)=1$ in later steps (staged scheduler in Table~\ref{tab:schedule_ablate}. This design effectively updates only the LLM parameters in the early phase and only the DiT parameters in the later phase.

\begin{figure}[!ht]
    \centering
    \includegraphics[width=0.8\linewidth]{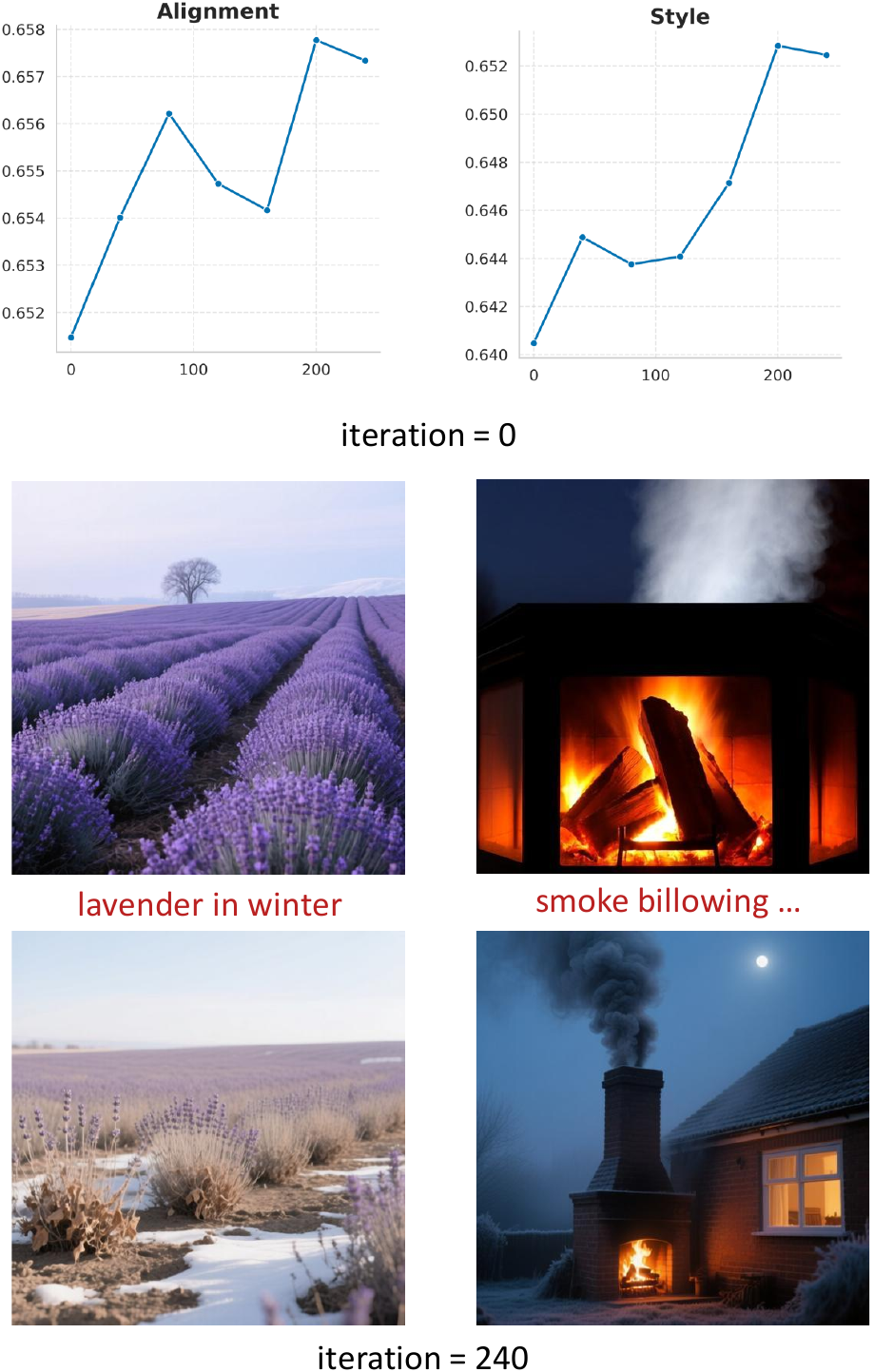}
    \caption{Evolution of alignment and style scores during training with the new scheduler, where $\beta_1(\tau) = \beta_2(\tau) = 0.5$, and samples demonstrating how DiT training improves both semantic alignment and visual quality over time.}
    \label{fig:new_scheduler}
\end{figure}

We compare the effectiveness of different schedulers in Table~\ref{tab:schedule_ablate} on the T2I-ReasonBench benchmark. The results show that the balanced scheduler consistently outperforms the staged scheduler. We hypothesize that this is because joint optimization enables tighter coordination between prompt refinement and visual rendering. In Figure~\ref{fig:new_scheduler}, we visualize the evolution of the rewards during training, as well as samples before and after the \textit{Dual}-GRPO under this new staged scheduler.

\begin{figure}[h]
    \centering
    \includegraphics[width=0.78\linewidth]{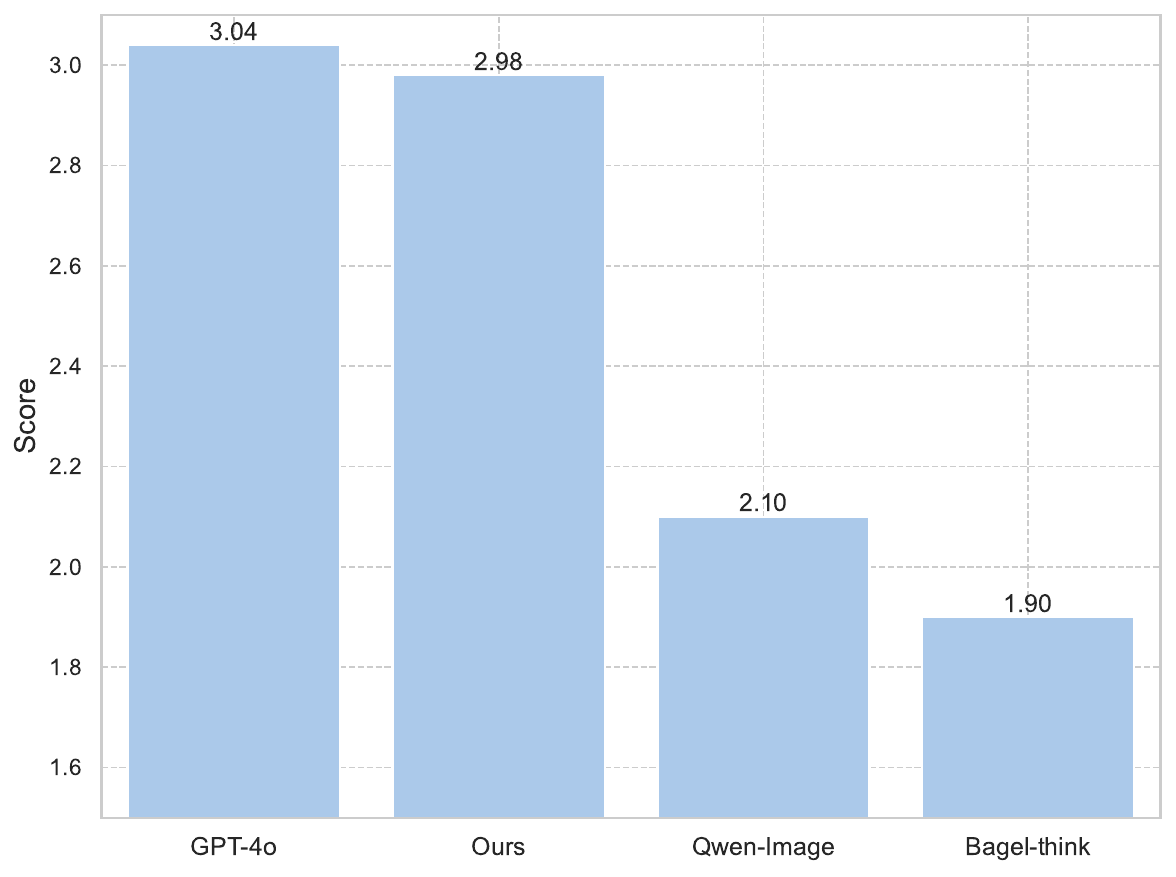}
    \caption{Comparison of scores on user study. Our model performs closely to GPT-4o and clearly surpasses Qwen-Image, highlighting its strong ability in challenging real-world T2I scenarios.}
    \label{fig:score}
\end{figure}

    
\begin{figure*}[t]
    \centering
    \includegraphics[width=0.84\linewidth]{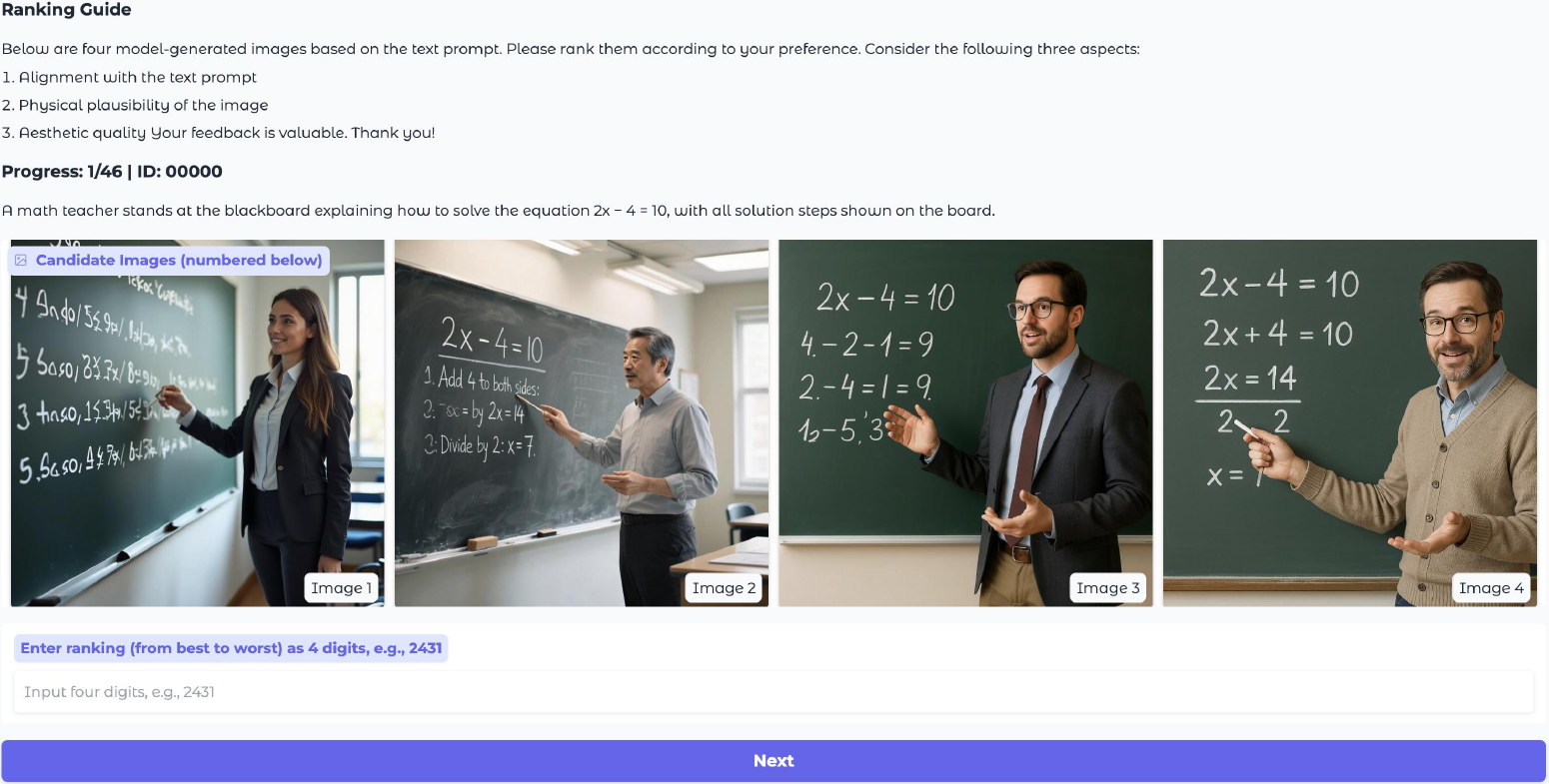}
    \caption{User interface for our human evaluation study. Images 1–4 correspond to Bagel-think, our model, vanilla Qwen-Image, and GPT-4o, respectively. Notably, our model provides the most comprehensive deduction for this math problem, uniquely arriving at the correct final answer, thereby demonstrating its superior reasoning ability and practical effectiveness in real-world tasks.}
    \label{fig:ranl_ui}
\end{figure*}

\begin{table*}[!ht]
    \centering
    \caption{Detailed scores on RISEBench. Numbers in bold indicate the highest score among open-source models.}
    \begin{tabular}{lcccccc}
    \toprule
    \multicolumn{1}{c}{Model} & Type                                                  & Temporal & Causal & Spatial & Logical & Overall $\uparrow$ \\ \midrule
    Gemini-2.5-Flash-Image    &  \textit{proprietary}                                                     & 25.9     & 47.8   & 37      & 18.8    & 32.8               \\
    GPT-Image-1               &   \textit{proprietary}                                                    & 34.1     & 32.2   & 37      & 10.6    & 28.9               \\
    Gemini-2.0-Flash-exp      &   \textit{proprietary}                                                     & 8.2      & 15.5   & 23      & 4.7     & 13.3               \\
    Seedream-4.0              & \textit{proprietary}                         & 12.9     & 12.2   & 11      & 7.1     & 10.8               \\
    BAGEL                     &       \textit{unified}                        & 2.4      & 5.6    & 14      & 1.2     & 6.1                \\
    BAGEL (w/ CoT)            &           \textit{unified}                    & 5.9      & 17.8   & 21      & 1.2     & 11.9               \\
    FLUX.1-Kontext        & \textit{diffusion} & 2.3      & 5.5    & 13.0      & 1.2     & 5.8      
    \\ 
    Qwen-Image-Edit           &        \textit{diffusion}                      & 4.7      & 10.0     & 17.0      & 2.4     & 8.9                \\
    \rowcolor[HTML]{ECF4FF} 
    Ours (w/o GRPO)                      & \textit{diffusion}                 &   \textbf{22.3}  &      27.8  &      21.0  &      9.4   &     20.2                       \\
    \rowcolor[HTML]{ECF4FF} 
    Ours                      & \cellcolor[HTML]{ECF4FF}   \textit{diffusion}                 &    20.0  &      \textbf{28.9}  &      \textbf{15.3}  &      \textbf{30.0}   &     \textbf{23.9}                      \\
    \bottomrule
    \end{tabular}
    \label{tab:append_rise}
\end{table*}

\section{Detailed Results on Image Editing Task}
\label{append:detailed_edit}
In this section, we provide detailed scores on the image editing task. We compare our model with several state-of-the-art image editing models over the RISE~\cite{zhao2025envisioning} benchmark. The results are shown in \ref{tab:append_rise}. It can be seen that training under our \textit{think-then-generate} paradigm significantly outperforms the vanilla Qwen-Image-Edit and even surpasses strong proprietary models like Seedream-4.0. These results demonstrate that incorporating reasoning into the generation process substantially improves both accuracy and visual quality of edited images.

\section{User Study}
In this section, we conduct a user study on T2I generation in challenging real-world scenarios. We select 46 prompts and compare our model against strong baselines, including vanilla Qwen-Image~\cite{wu2025qwen}, Bagel-think~\cite{deng2025emerging}, and GPT-4o~\cite{hurst2024gpt}.
Users are instructed to rank the model outputs according to \textit{semantic alignment and conceptual understanding}, \textit{visual realism and coherence}, and \textit{aesthetic quality}. Our preference-collection interface is shown in Figure~\ref{fig:ranl_ui}.

For evaluation, we aggregate prompt-level rankings. For each given prompt $i$, the model receives a reward $s_i = 5-r_i$, where $r_i \in \{1,2,3,4\}$ denotes its rank among the four models. The average score is computed by adding the rewards and averaging all prompts (Figure~\ref{fig:score}). Results show that our model performs closely to GPT-4o and clearly surpasses Qwen-Image, highlighting its strong ability in challenging real-world T2I scenarios. Notably, as shown in Figure~\ref{fig:ranl_ui}, our model produces the most comprehensive reasoning for math-related problems and uniquely arrives at the correct final answer, highlighting its superior deductive capabilities and practical effectiveness in real-world tasks.


\section{More Demos}
\begin{figure*}[t]
    \centering
    \includegraphics[width=0.94\linewidth]{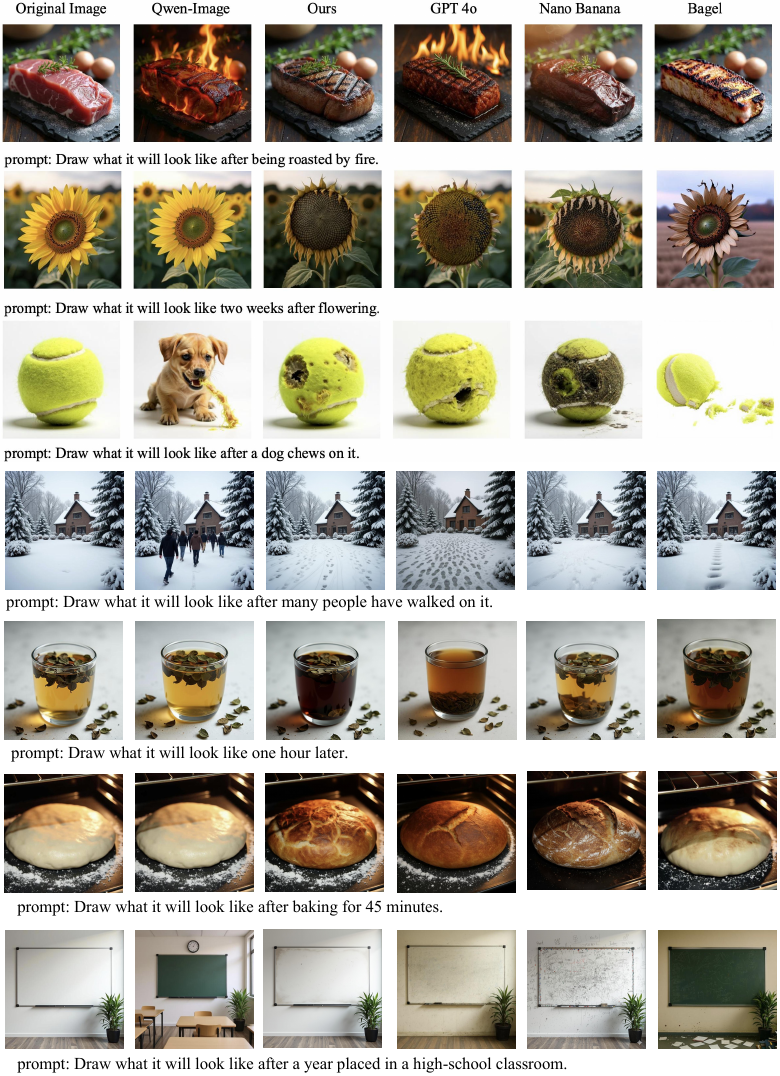}
    \caption{More demos for image editing task.}
    \label{fig:append_edit_demo}
\end{figure*}

\begin{figure*}[t]
    \centering
    \includegraphics[width=0.97\linewidth]{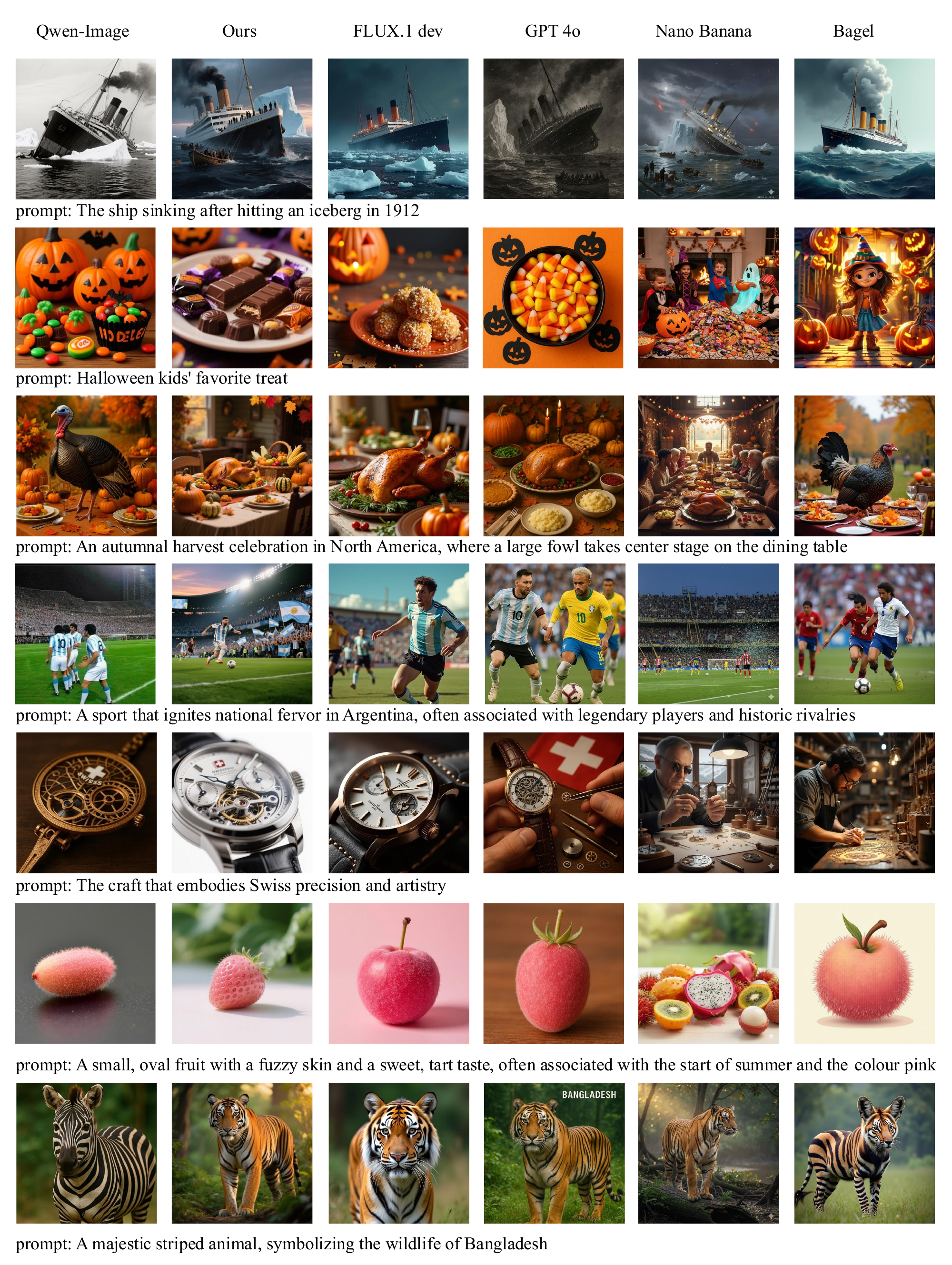}
    \caption{More demos for the T2I task.}
    \label{fig:demo_t2i_sup1}
\end{figure*}

We provide more demos for the T2I task in Figure~\ref{fig:demo_t2i_sup1}, and additional examples for image editing in Figure~\ref{fig:append_edit_demo}. These results further demonstrate that our model delivers stronger conceptual instruction understanding and alignment, more robust appearance consistency across generated content, and improved overall visual plausibility and coherence compared to existing baselines.


